\documentclass{article}

\usepackage[T1]{fontenc}
\usepackage{microtype}
\usepackage{graphicx}
\usepackage{booktabs}
\usepackage{amsmath,amssymb}
\usepackage{hyperref}

\usepackage[preprint]{icml2026}

\icmltitlerunning{Mathematical Reasoning in Turkish LLMs}
\icmlkeywords{LLM reasoning, Turkish NLP, fine-tuning, Tree of Thoughts, model merging, tree search}

\begin{document}

\twocolumn[
\icmltitle{Enhancing Mathematical Reasoning in Turkish Large Language Models with Search-Based Inference and Model Merging}

\begin{icmlauthorlist}
\icmlauthor{Ahmed Zeer}{ytu}
\icmlauthor{Osama SHBIB}{ytu}
\icmlauthor{Muzaffer Kaan Yuce}{ytu}
\icmlauthor{Himmet Toprak Kesgin}{ytu}
\icmlauthor{Mehmet Fatih Amasyali}{ytu}
\end{icmlauthorlist}

\icmlaffiliation{ytu}{Yildiz Technical University, Istanbul, T\"urkiye}
\icmlcorrespondingauthor{Ahmed Zeer}{ahmed.zeer@std.yildiz.edu.tr}
\vskip 0.3in
]

\printAffiliationsAndNotice{}
\hypersetup{
  pdftitle={Enhancing Mathematical Reasoning in Turkish Large Language Models with Search-Based Inference and Model Merging},
  pdfauthor={Ahmed Zeer, Osama SHBIB, Muzaffer Kaan Yuce, Himmet Toprak Kesgin, Mehmet Fatih Amasyali},
  pdfsubject={Mathematical reasoning in Turkish large language models}
}

\begin{abstract}
We improve Turkish mathematical reasoning in Large Language Models by combining LoRA fine-tuning, weight-space model merging, and search-based inference. Using Turkish-Llama-8B-DPO with Turkish translations of MetaMath, NuminaMath, and TigerLab, we find data quality beats scale (Mix-30k best among fine-tuned; large MetaMath underperforms). Linear merging delivers the largest gains (GSM8K up to 0.63), while Ties is modest and Breadcrumbs/Slerp degrade. Search with MCTS and Tree of Thoughts improves mean LLM-judge scores for reasoning quality on a 0--100 scale. Limitations include machine translation, limited scale, and search stochasticity.

\end{abstract}

\section{Introduction}
Large Language Model (LLM) mathematical problem solving is challenging, requiring structured reasoning and precise calculations. One choice to enhance the model's reasoning capabilities is by training it with domain-specific datasets. The other is to give the model a chance to reflect on its own thoughts and have a couple of structured decoding iterations. The earlier method requires training and data curation when the latter does not.

Low-to-mid-resource languages lag behind due to data scarcity. This disparity necessitates approaches that maximize the utility of available translated data and pretrained weights. Consequently, our work explores how algorithmic interventions---specifically inference-time search and weight-space merging---can bridge this performance gap without requiring massive-scale pre-training from scratch.

To achieve this, we utilize one of the best performing Turkish LLMs, \textit{Turkish-Llama-8b-DPO-v0.1} \citep{Kesgin2024OptimizingLL} (COSMOS-DPO) according to the \textit{OpenLLMTurkishLeaderboard} \citep{openllm-Turkish-leaderboard} on \textit{Huggingface}. Our study investigates the following key aspects:
\begin{enumerate}
    \item Assessing the impact of different math datasets on fine-tuning performance in Turkish.
    \item Evaluating the effect of merging fine-tuned models with earlier versions to enhance reasoning abilities.
    \item Comparing search algorithms, including Monte Carlo Tree Search (MCTS) and Tree of Thoughts (ToT) \citep{Yao2023TreeOT}, for mathematical reasoning in Turkish.
    \item Design a modular Reward-Judge-Enhance pipeline to simulate self-reflection and enhance model's reasoning capabilities.
    \item Evaluating the influence of each module in the Reward-Judge-Enhance pipeline for determining roles and importance.
\end{enumerate}

Through this work, we aim to contribute to the development of mathematical reasoning in Turkish LLMs and provide insights into optimizing fine-tuning, search-based inference, weight merging strategies.

\section{Related Work}

\subsection{LLM Mathematical Reasoning and Search}

Giving the model a chance to think more and evaluate current paths, instead of answering the current questions directly, is one of the recent methods to enhance the LLM's output. While some methods structure possible solutions and paths in a hierarchical way and perform a search to possibly find the best solution \citep{Yao2023TreeOT} \citep{chen2024alphamath} \citep{guan2025rstar}, other methods aligns the model, by either training or few-shot prompting, to think step-by-step \citep{Wei2022ChainOT}.

\subsection{Model Merging Techniques}
Model merging has emerged as a cost-effective alternative to continuous pre-training. Techniques range from simple linear averaging \citep{wortsman2022modelsoups} to interference-aware methods like TIES \citep{Yadav2023TIESMergingRI} and DARE \citep{Yu2023LanguageMA}. However, this study shows that complex merging heuristics often fail to outperform simple ones for reasoning tasks.

\section{Methodology}
\subsection{Fine-Tuning with Math Datasets}

\begin{figure*}[!t]
    \centering
    \includegraphics[width=0.60\textwidth]{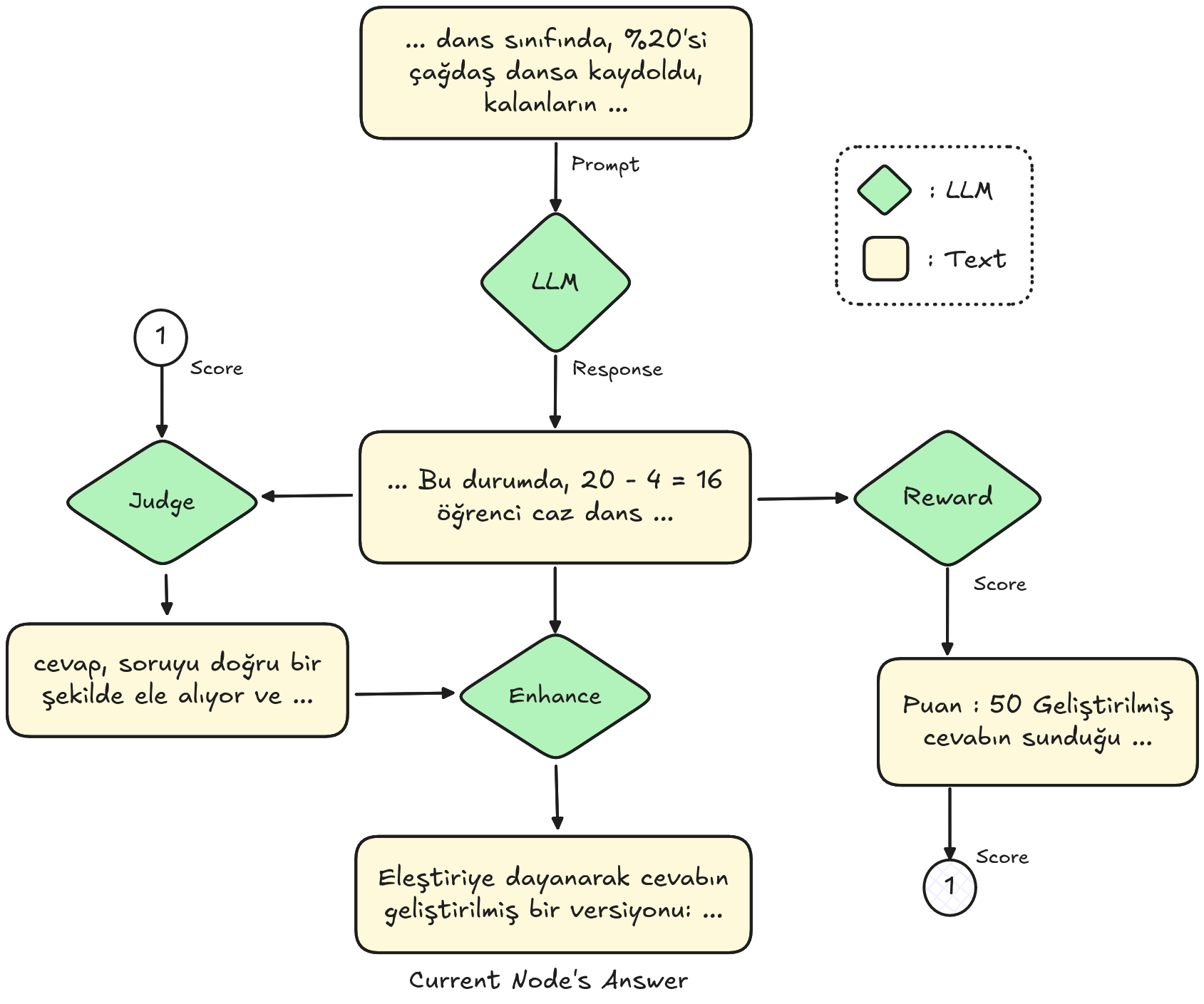} 
    \caption{Reward-Judge-Enhancement Pipeline.}
    \label{fig:harezmi-pipeline-dikey}
\end{figure*}

\begin{figure*}[!t]
    \centering
    \includegraphics[width=1.0\textwidth]{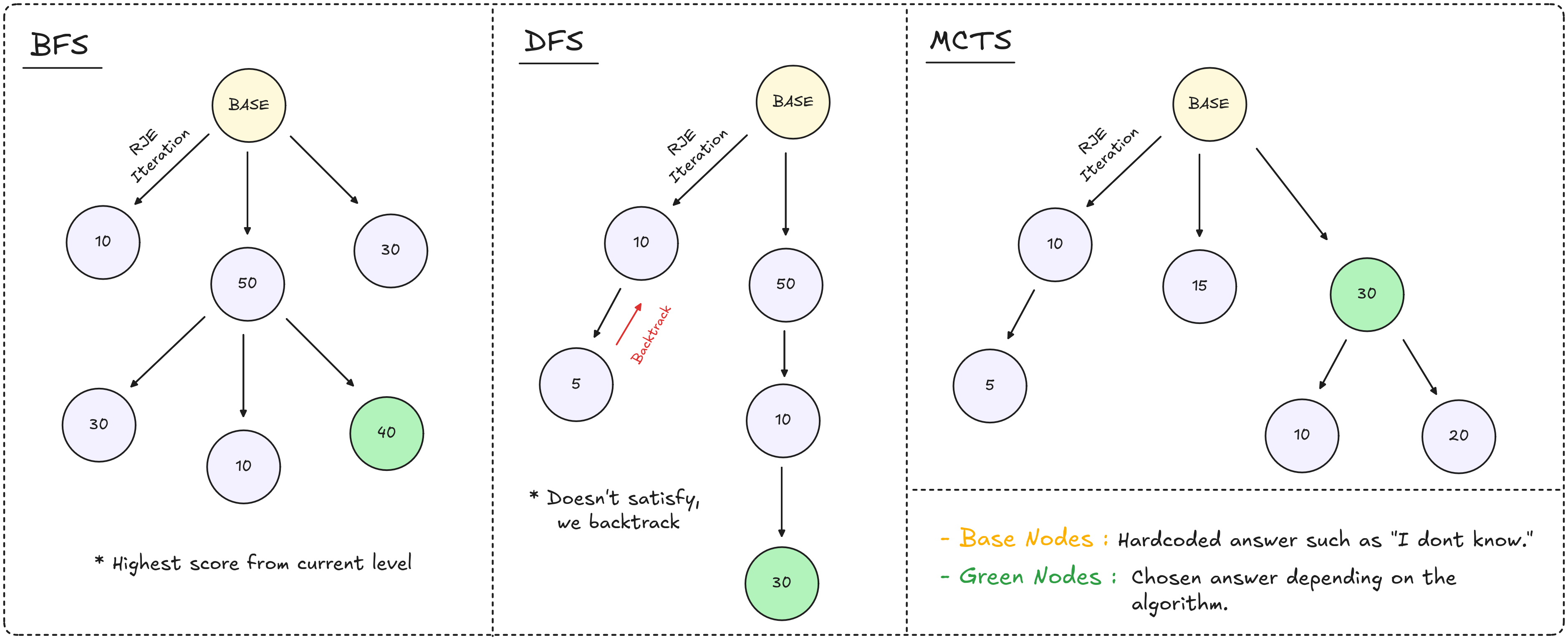} 
    \caption{Algorithm based tree of thought generation.}
    \label{fig:tot-harezmi}
\end{figure*}

Supervised fine-tuning on mathematical datasets has shown to enhance the mathematical reasoning capabilities of LLMs \citep{Cobbe2021TrainingVT}. In this study, we employ Low-Rank Adaptation (LoRA) \citep{Hu2021LoRALA} to efficiently fine-tune LLMs on three mathematical datasets: \textbf{MetaMath} \citep{Yu2023MetaMathBY}, \textbf{NuminaMath} \citep{numina_math_datasets}, and \textbf{TigerLab-MathInst} \citep{yue2023mammoth}. Translation process has been done using proprietary models like that of DeepL.

Each dataset provides distinct mathematical challenges and reasoning structures. The impact of fine-tuning on mathematical performance is evaluated through standard benchmarks. To test the performance of the datasets only 50k samples were chosen from each dataset. Also \textbf{Mix} dataset where created by sampling random number of questions from all of the datasets. Different mixes increases diversity and prevent overfitting to a single reasoning style. 

\subsection{Model Merging}
We experimented with various merging techniques to create new models using different combinations of fine-tuned versions. The techniques employed include \textbf{Linear} \citep{Zhou2025MixupMM}, which averages the model weights in a straightforward manner; \textbf{Ties} \citep{Yadav2023TIESMergingRI}, which selects parameters based on confidence thresholds; \textbf{Breadcrumbs} \citep{Davari2023ModelBS}, which merges models by tracking gradient updates across fine-tuned versions; and \textbf{Slerp} (Spherical Linear Interpolation) \citep{Jang2024SphericalLI}, which interpolates between model weight spaces to achieve smoother transitions. By applying these approaches, we aimed to enhance the mathematical reasoning capabilities of our models and systematically evaluate the impact of merging on overall performance.

\subsection{Search Algorithms}
Search algorithms offer a significant advantage in improving LLM performance by guiding the reasoning process towards more accurate solutions. In this study, we explored three key search-based approaches: \textbf{Monte Carlo Tree Search (MCTS)} and \textbf{Tree of Thoughts (ToT)} \citep{Yao2023TreeOT}. Using these abstract search methods for searching in a tree of thoughts is highly implementation dependent. We elaborate algorithms' expected behavior in Figure ~\ref{fig:tot-harezmi}.

\begin{algorithm}[!t]
\footnotesize
\caption{Breadth-First Search (ToT-BFS)}\label{alg:bfs}
\begin{algorithmic}[1]
\REQUIRE Root Node $root$, Max Depth $D$, Max Iterations $N$
\STATE $Queue \gets [(root, 0)]$
\STATE $best\_node \gets root$
\STATE $iter \gets 0$
\WHILE{$Queue \neq \emptyset$ \textbf{and} $iter < N$}
    \STATE $(node, depth) \gets Queue.popleft()$
    \IF{$node.rating$ is None}
        \STATE $Evaluate(node)$
    \ENDIF
    \IF{$node.rating > best\_node.rating$}
        \STATE $best\_node \gets node$
    \ENDIF
    \IF{$depth < D$}
        \IF{$node.children$ is empty}
            \STATE $Expand(node)$ \COMMENT{Generates children via RJE}
        \ENDIF
        \FOR{$child$ \textbf{in} $node.children$}
            \STATE $Queue.append((child, depth + 1))$
        \ENDFOR
    \ENDIF
    \STATE $iter \gets iter + 1$
\ENDWHILE
\STATE \textbf{return} $best\_node$
\end{algorithmic}
\end{algorithm}

Searching for possible solutions in each level is Breadth First Search algorithm's job (Algorithm~\ref{alg:bfs}). On the other hand, Depth First Search rolls the current node until it reaches maximum and backtracks to search for other possibilities (Algorithm~\ref{alg:dfs}).

\begin{algorithm}[!t]
\footnotesize
\caption{Depth-First Search (ToT-DFS)}\label{alg:dfs}
\begin{algorithmic}[1]
\REQUIRE Root Node $root$, Max Depth $D$, Max Iterations $N$
\STATE $Stack \gets [(root, 0)]$
\STATE $best\_node \gets root$
\STATE $iter \gets 0$
\WHILE{$Stack \neq \emptyset$ \textbf{and} $iter < N$}
    \STATE $(node, depth) \gets Stack.pop()$
    \IF{$node.rating$ is None}
        \STATE $Evaluate(node)$
    \ENDIF
    \IF{$node.rating > best\_node.rating$}
        \STATE $best\_node \gets node$
    \ENDIF
    \IF{$depth < D$}
        \IF{$node.children$ is empty}
            \STATE $Expand(node)$ \COMMENT{Generates children via RJE}
        \ENDIF
        \FOR{$child$ \textbf{in} $node.children$}
            \STATE $Stack.append((child, depth + 1))$
        \ENDFOR
    \ENDIF
    \STATE $iter \gets iter + 1$
\ENDWHILE
\STATE \textbf{return} $best\_node$
\end{algorithmic}
\end{algorithm}

MCTS, tries to balance between former methods and chooses the maximum node globally (Algorithm~\ref{alg:mcts}).

\begin{algorithm*}[!t]
\small
\caption{Monte Carlo Tree Search with RJE}\label{alg:mcts}
\begin{algorithmic}[1]
\REQUIRE Question $Q$, Seed Answers $S$, Iterations $N$
\STATE $root \gets Node(Q, random.choice(S))$
\FOR{$i \gets 1$ \textbf{to} $N$}
    \STATE $node \gets root$
    \COMMENT{Selection}
    \WHILE{$node$ is fully expanded \textbf{and} $node.children \neq \emptyset$}
        \STATE $node \gets BestChild(node)$ \COMMENT{Using UCB formula}
    \ENDWHILE
    
    \COMMENT{Expansion with Reward-Judge-Enhance}
    \IF{$len(node.children) < MaxChildren$}
        \STATE $child \gets Node(Q, node.answer, parent=node)$
        \STATE $critique \gets CritiqueClient(node.question, child.answer)$
        \STATE $enhanced \gets EnhanceClient(node.question, child.answer, critique)$
        \STATE $child.answer \gets enhanced$
        \STATE $node.children.append(child)$
        \STATE $node \gets child$
    \ENDIF
    
    \COMMENT{Simulation}
    \STATE $reward \gets RewardClient(node.question, node.answer)$
    
    \COMMENT{Backpropagation}
    \WHILE{$node \neq \textbf{null}$}
        \STATE $node.visits \gets node.visits + 1$
        \STATE $node.value \gets node.value + reward$
        \STATE $node \gets node.parent$
    \ENDWHILE
\ENDFOR
\STATE \textbf{return} $MostVisitedChild(root)$
\end{algorithmic}
\end{algorithm*}

Each node is not only created by the answer it holds only. The answer is passed through the Reward-Judge-Enhance (RJE) pipeline and corresponding metadata is generated. The initial answer gets a reward then immediately judged by a criticizing agent, can be the model itself, given the criticism and the current answer an enhancement agent is used to further refine it. The procedure is illustrated in Figure~\ref{fig:harezmi-pipeline-dikey}.

To systematically assess these search algorithms, we developed four client configurations (Table~\ref{tab:search_clients}). Notably, \textbf{Client-1} relies on GPT \citep{Brown2020LanguageMA} for reward assignment and judgment, but importantly, it uses the local model for the \textbf{Enhance} (Self-Correction) step. This distinguishes our setup from pure distillation; the reasoning generation remains local, while the evaluation signal is external.

For evaluation, we tested these clients on a 100 questions subset of the Turkish version of \textbf{GSM8K}, specifically \texttt{gsm8k\_tr-v0.2} \citep{gsm8k_tr-v0.2}.

\begin{figure*}[!t]
    \centering
    \includegraphics[width=0.78\textwidth]{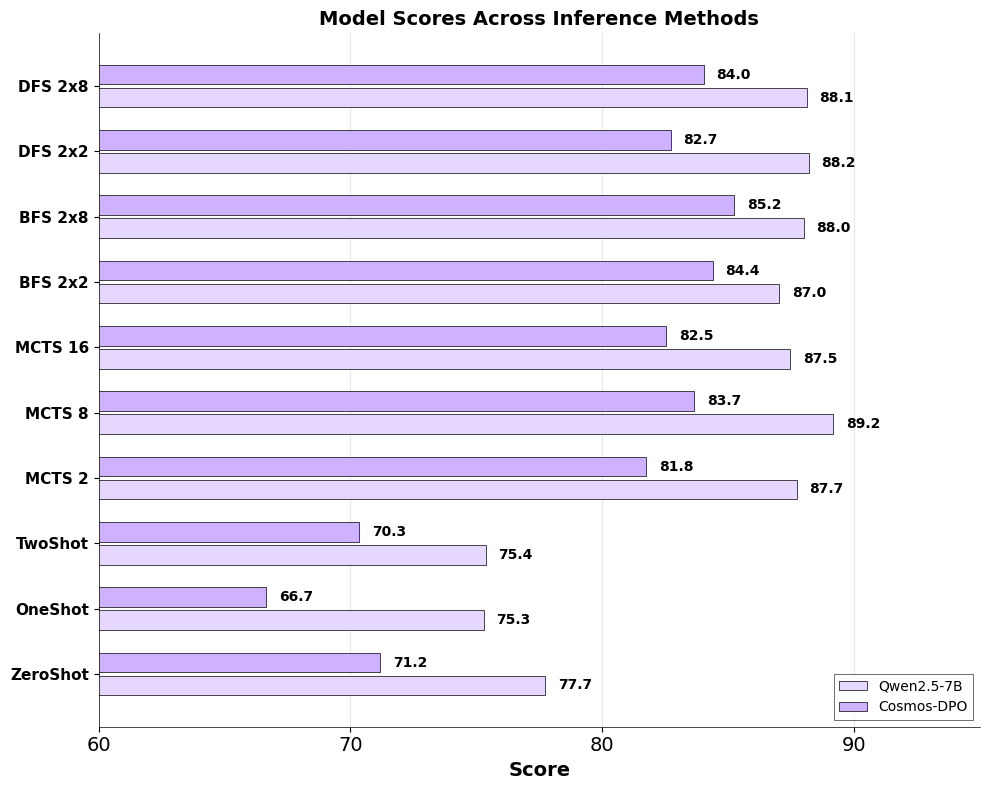} 
    \caption{Mean LLM-judge scores (0--100) across inference methods: Qwen2.5-7B vs Cosmos-DPO.}
    \label{fig:performance-comparison}
\end{figure*}

\begin{figure}[!t]
    \centering
    \includegraphics[width=\linewidth]{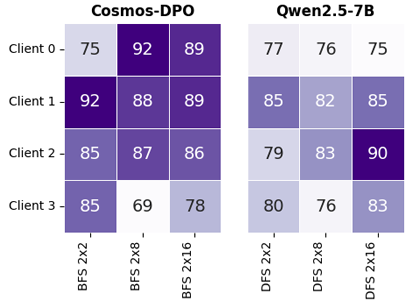} 
    \caption{Mean LLM-judge scores (0--100) by method and client.}
    \label{fig:clients_viz}
\end{figure}

\begin{figure*}[!t]
    \centering
    \includegraphics[width=0.80\textwidth]{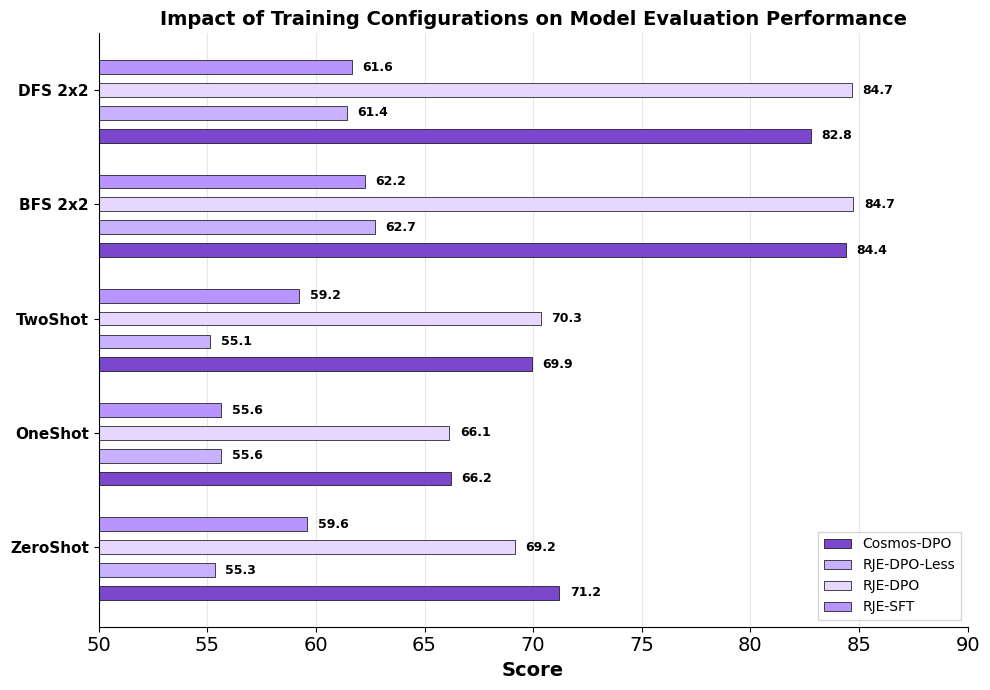} 
    \caption{Mean LLM-judge scores (0--100) for Cosmos-DPO before and after training.}
    \label{fig:cosmos-dpo-train}
\end{figure*}

\begin{table*}[!t]
    \caption{Performance comparison across different datasets and methods. The highest values in GSM8K and Average are highlighted.}
    \label{tab:math_finetune_results}
    \centering
    \small
    \setlength{\tabcolsep}{4pt}
    \begin{tabular}{llllll | l}
        \toprule
        & ARC & GSM8K & Hellaswag & TruthfulQA & Winogrande & \textbf{Average} \\
        \midrule
        Mix-50K & 0.48 & 0.55 & 0.50 & 0.53 & 0.57 &  0.52 \\
        Mix-40K & 0.48 & 0.55 & 0.50 & 0.53 & 0.57 &  0.52 \\
        Mix-30K & 0.49 & \textbf{0.57} & 0.51 & 0.54 & 0.56 &  0.53 \\
        Mix-20K & 0.47 & 0.56 & 0.51 & 0.54 & 0.56 &  0.52 \\
        Metamath-50K & 0.47 & 0.55 & 0.51 & 0.52 & 0.57 &  0.52 \\
        Numina-50K & 0.47 & 0.54 & 0.50 & 0.52 & 0.57 & 0.52 \\
        TigerLab-MathInst-50K & 0.48 & 0.47 & 0.51 & 0.53 & 0.56 & 0.51 \\
        Metamath-300K & - & 0.49 & - & - & - & - \\
        \midrule
        \textit{Baseline (Cosmos-DPO)} & \textit{0.51} & \textit{0.57} & \textit{0.52} & \textit{0.57} & \textit{0.57} & \textbf{0.55} \\
        \bottomrule
    \end{tabular}
\end{table*}

\begin{table*}[!t]
  \caption{Performance of different merging techniques on GSM8K.}
  \label{tab:merged_models}
  \centering
    \small
    \setlength{\tabcolsep}{4pt}
  \begin{tabular}{llc}
    \toprule
    \textbf{Merged Models} & \textbf{Method} & \textbf{GSM8K} \\
    \midrule
    DPO, Metamath-50K        & Linear                        & 0.62 \\
    DPO, Metamath-50K        & Ties                          & 0.57 \\
    DPO, Numina-50K          & Ties                          & 0.58 \\
    DPO, Metamath-50K        & Breadcrumbs                   & 0.54 \\
    DPO, Metamath-50K        & Slerp                         & 0.25 \\
    Metamath-50K, DPO        & Slerp                         & 0.26 \\
    DPO, Metamath, Numina    & Linear - Same Time            & 0.60 \\
    DPO, Metamath, Numina    & Linear - Binary               & 0.61 \\
    DPO, Metamath-50K        & Evolve - Linear (24 Steps)    & 0.62 \\
    DPO, Metamath-300K       & Linear                        & \textbf{0.63} \\
    \textit{Baseline (Cosmos DPO)} & -                         & 0.57 \\
    \bottomrule
  \end{tabular}
\end{table*}

\begin{table}[!t]
    \caption{Inference Time (Seconds) for 100 Questions on A100 80GB VRAM.}
    \label{tab:compute_cost}
    \centering
    \small
    \setlength{\tabcolsep}{4pt}
    \begin{tabular}{lr}
        \toprule
        \textbf{Method} & \textbf{Time (s)} \\
        \midrule
        Zero-shot & 23.72 \\
        One-shot & 32.75 \\
        Two-shot & 37.66 \\
        MCTS 2 Iterations & 224.56 \\
        ToT BFS 2x2 & 550.45 \\
        MCTS 16 Iterations & 2035.28 \\
        ToT DFS 16x16 & 12087.72 \\
        \bottomrule
    \end{tabular}
\end{table}

\begin{table}[!t]
    \caption{Client configurations for search algorithm evaluation.}
    \label{tab:search_clients}
    \centering
    \small
    \setlength{\tabcolsep}{4pt}
    \begin{tabular}{clll}
        \toprule
        \textbf{Client} & \textbf{Reward} & \textbf{Judge}  & \textbf{Enhance} \\
        \midrule
        Client-0 & Self-Reward & Self-Judge & Self-Enhance \\
        Client-1 & GPT & GPT & Self-Enhance \\
        Client-2 & Self-Reward & GPT & Self-Enhance \\
        Client-3 & GPT & Self-Judge & Self-Enhance \\
        \bottomrule
    \end{tabular}
\end{table}

\textbf{Search Specifications:}

For ToT search methods, we adopt the convention \textbf{CxD}, where $C$ is child nodes per expansion and $D$ is maximum depth. For MCTS, we use \textbf{MCTS X}, where $X$ indicates the number of iterations. MCTS implementation initializes with generic seed answers ("I don't know"). Upper Confidence Bound exploration weight of 1.41 and limit expansion to a maximum of 2 children per node is used to balance breadth and depth.

\section{Results and Discussion}
\subsection{Fine-Tuning}

Table \ref{tab:math_finetune_results} presents a comparative analysis of various fine-tuning approaches on multiple benchmarks. The evaluation is done using lm-eval-harness suite \citep{eval-harness}, which contains Turkish datasets also. The primary focus is on assessing the impact of different dataset configurations on GSM8K performance. Even with different sizes and diversity mixtures Finetuning driven approach couldn't pass the baseline as seen in the table. Still, as we will see in the next subsection, weight merging these models with the original increases benchmark scores.

\subsection{Model Merging}
Table~\ref{tab:merged_models} indicates that linear merging delivers the strongest GSM8K gains, with several variants reaching 0.62--0.63. The best model, \emph{DPO, Metamath-300K (Linear)}, attains 0.63. Iterative refinement via \emph{Evolve -- Linear} over 24 steps achieves 0.62. In contrast, Ties merging yields mixed results (0.57--0.58).
We hypothesize that Ties merging, by aggressively pruning parameters with conflicting signs, may inadvertently discard subtle but necessary redundancies. Other techniques, such as Breadcrumbs and Slerp, struggled to improve results.

The success of Linear merging over Slerp supports the hypothesis of 'linear mode connectivity' among models fine-tuned from the same base. Since all our variants utilized the same pre-trained backbone (Cosmos-DPO), their weight updates likely occurred within a constrained subspace.

\subsection{Search Algorithms and Compute Costs}
Our experiments show that math-focused search-based inference improves LLM-judge scores for Turkish mathematical reasoning (Fig.~\ref{fig:performance-comparison}). Both \textbf{Qwen2.5-7B} and \textbf{Cosmos-DPO} surpass zero-/few-shot baselines on this metric once search is added. Fine-tuning and merging results in Tables~\ref{tab:math_finetune_results} and~\ref{tab:merged_models} use the evaluation harness \citep{eval-harness}. For the inference and RJE training comparisons in Figs.~\ref{fig:performance-comparison},~\ref{fig:clients_viz}, and~\ref{fig:cosmos-dpo-train}, we instead evaluate response quality using an LLM judge. After inference, the evaluator supplies the question, generated response, and reference answer to \texttt{gpt-4o-mini} at temperature 0. The judge assigns a graded score from 0 to 100 based on correctness, logical consistency, and reasoning quality, allowing partial credit for imperfect solutions. We report the arithmetic mean of these scores across evaluated questions; averaging graded scores explains decimal values such as 84.4. These values are mean judge scores, not percentages of questions answered correctly, and are not directly comparable to the harness benchmark accuracies.

These improvements in mean judge score come at a significant computational cost. Table \ref{tab:compute_cost} details the inference timing on an A100 80GB GPU. While standard Zero-shot inference takes approximately 24 seconds for 100 questions, advanced search methods like ToT (16x16 DFS) can take over 12,000 seconds---a 500x increase in latency. This computational bottleneck explains why our search evaluations are limited to relatively small subsets.

Despite the costs, our results highlight a distinct 'efficiency sweet spot' at the BFS 2x2 configuration. As illustrated in Figure \ref{fig:clients_viz}, this lower-depth search achieves mean judge scores of 92 for Cosmos-DPO and 85 for Qwen2.5-7B on Client-1, without the exponential compute explosion seen in deeper trees.

These findings inspired a key question: \textit{Can we improve the model's ability to judge, refine, and reward its own responses?} We refer to the pipeline using the acronym RJE (Reward--Judge--Enhance). As shown in Fig.~\ref{fig:cosmos-dpo-train}, none of these training methods yield a significant performance improvement.

Taken together, these experiments show that search improves mean LLM-judge scores, while merging improves benchmark accuracy. Because the evaluation procedures differ, these results do not establish a direct ranking of search and merging.

\section{Conclusion}
We studied LoRA fine-tuning, linear model merging, and search-based inference to improve math reasoning in Turkish LLMs. Linear merges consistently helped, with a best GSM8K score of 0.63; optimized MCTS and Tree of Thoughts improved mean LLM-judge scores for response quality, though at significant computational cost (up to 500x latency). Still, the results shows that search-based inference techniques seems to saturate after a certain depth.

We relied on the `gsm8k\_tr-v0.2` dataset, which was generated via GPT-4; while high quality, it may contain specific translation artifacts. Our search evaluations were restricted to 100-question subsets due to the massive inference costs (detailed in Table \ref{tab:compute_cost}). We did not run human evaluations on the translations.

\section*{Acknowledgments} This study was supported by the Scientific and Technological Research Council of Türkiye (TUBITAK) Grant No: 124E055.

\section*{Disclosure of Interests}
The authors have no competing interests to declare that are
relevant to the content of this article.

\nocite{*}
\bibliographystyle{icml2026}
\bibliography{references}

\end{document}